\documentclass[preprint,12pt]{elsarticle}

\usepackage{amssymb, xcolor, hyperref, bm, subcaption, caption, multirow}
\usepackage{amsmath, cleveref, soul, xcolor, placeins}

\newcommand{\mli}[1]{\mathit{#1}}

\journal{Information Sciences}

\begin{document}

\begin{frontmatter}



\title{Assessing Fidelity in XAI post-hoc techniques: A Comparative Study with Ground Truth Explanations Datasets}

\author[1,2]{Miquel Mir\'{o}-Nicolau}
\ead{miquel.miro@uib.es}
\author[1,2]{Antoni Jaume-i-Cap\'{o}}
\ead{antoni.jaume@uib.es}
\author[1,2]{Gabriel Moy\`{a}-Alcover} \corref{cor1}
\ead{gabriel.moya@uib.es}

\cortext[cor1]{Corresponding author}
\address[1]{UGiVIA Research Group, University of the Balearic Islands, Dpt. of Mathematics and Computer Science, 07122 Palma (Spain)}
\address[2]{Laboratory for Artificial Intelligence Applications (LAIA@UIB), University of the Balearic Islands, Dpt. of Mathematics and Computer Science, 07122 Palma (Spain)}



            
\begin{abstract}
The evaluation of the fidelity of eXplainable Artificial Intelligence (XAI) methods to their underlying models is a challenging task, primarily due to the absence of a ground truth for explanations. However, assessing fidelity is a necessary step for ensuring a correct XAI methodology. In this study, we conduct a fair and objective comparison of the current state-of-the-art XAI methods by introducing three novel image datasets with reliable ground truth for explanations. The primary objective of this comparison is to identify methods with low fidelity and eliminate them from further research, thereby promoting the development of more trustworthy and effective XAI techniques. Our results demonstrate that XAI methods based on the backpropagation of output information to input yield higher accuracy and reliability compared to methods relying on sensitivity analysis or Class Activation Maps (CAM). However, the backpropagation method tends to generate more noisy saliency maps. These findings have significant implications for the advancement of XAI methods, enabling the elimination of erroneous explanations and fostering the development of more robust and reliable XAI.

\end{abstract}


\end{frontmatter}


\section{Introduction}

The usage of Deep Learning models has become the gold standard for solving most artificial intelligence problems, starting with the seminal work of Krizhevsky et al. \cite{krizhevsky2012imagenet}, due to their much better results in comparison to other approaches. These better results are obtained thanks to an increase in complexity that at the same time turns these models into a black-box.

Due to the unawareness of the reasons behind the good results of these methods, its usage in sensitive field, such as medical practice, had been criticized \cite{adadi2018}. To address this issue, eXplainable Artificial Intelligence (XAI) emerged, aiming to shift to a more transparent AI \cite{adadi2018}. XAI methods enable users to gain insight into how a model arrives at its predictions, by providing explanations that can be understood and validated. By increasing the interpretability of Deep Learning models, XAI has the potential to unlock their full potential for a range of important applications as for example medical domain (\cite{eitel2019testing}, \cite{miro2022evaluating}, \cite{VanderVelden2021}). However, XAI is still a work in progress, with a lack of consensus and different approaches aiming to accomplish its basic goal to explain complex methods. 

Multiple authors had reviewed the XAI state-of-art (\cite{linardatos2020explainable}, \cite{adadi2018}, \cite{miro2022evaluating}, \cite{VanderVelden2021}). From these reviews, multiple conclusions can be obtained: first, interpretation methods can be classified either as local or global methods according to whether they aim to explain the whole logic of a model and follow the entire reasoning leading to all the different possible outcomes or to explain the reasons for a specific decision or single prediction \cite{adadi2018}; second, most XAI research aimed to develop local methods due to their inherent simplicity compared to global ones; third, saliency maps, visualizations that indicate the importance of each pixel, are the most used visualizations to explain image models; and, finally, that exists a large variety of XAI methods, and their explanations are not coherent among them.

The non-coherence between methods has been studied by Adebayo \emph{et al.}~\cite{adebayo2018sanity}, who proposed a methodology to solve it. In particular, they indicated the need to evaluate, in a machine-centric approach, the fidelity of the explanation. Fidelity, according to Tomsett \emph{et al.}~\cite{tomsett2020sanity} is how well the explanation agrees with the way the model actually works. Miller \cite{miller2019explanation} also indicated the need for a machine centric approach to measure fidelity and avoid human psychological biases. However, to calculate the fidelity exists a big limitation: the inability to have a ground truth of the real explanation. Taking all these facts in consideration, a set of methods emerged aiming to evaluate the fidelity of the explanations without a ground truth. The majority of the proposed measurements are based on some assumption about the relation between the explanation and the model. The basis of many of these methods is to expect a direct relation between the removal of important features, according to the explanation, and the overall performance of the predictive model \cite{bach2015pixel, Samek2017, Petsiuk2018, Alvarez-Melis2018, montavon2018methods, ancona2017towards, arya2019one, yeh2019fidelity, bhatt2020evaluating, rieger2020irof, rong2022consistent}. However, all these metrics tend to generate out-of-domain (OOD) samples \cite{qiu2021resisting, Gomez2022}, and this is the factor why these metrics are unreliable \cite{tomsett2020sanity}, 

To overcome the metric limitations, several authors have proposed generating synthetic datasets with ground-truth for the explanations. Cortez and Embrechts~\cite{CORTEZ20131} proposed a new XAI method and used a synthetic dataset, of 1000 tabular data samples, to measure its fidelity, with each of these 1000 samples containing 4 features and a synthetic label. The label was generated calculating a weighted sum of the dataset's features. An AI model was trained to regress this synthetic label from the data. The resultant model weights each feature of the dataset with the weights of the sum that generates the label, working as a ground truth for the explanations. The proposed dataset of Cortez and Embrechts~\cite{CORTEZ20131} is limited to tabular data. In our previous work\cite{miro2022evaluating}, we adapted and generalized the approach of Cortez \emph{et al.}~\cite{CORTEZ20131} to image data and for classification and regression tasks. We defined a synthetic dataset similar to Cortez \emph{et al.} with visual features of an input image, such as the number of times a visual pattern appears. However, we did not use it to compare any XAI methods. On the other hand, the generated datasets, with the goal to be easily verifiable, limit the possibility of OOD emergence. Arras \emph{et al.}~\cite{arras2020ground} proposed a methodology for generating a synthetic attribution dataset of images that only works for visual question answering (VQA) models. They proposed to ask questions depending only on one object from the image, only considering an explanation correct if the object of interest in the question was highlighted. Another limitation of this methodology is that they also did not consider the importance value of the explanations; instead, they only considered the position of the most important elements. Guidotti~\cite{guidotti_evaluating_2021} considered the black-box models trained on datasets with explanations ground truth to become transparent methods due to the availability of the real explanation. He proposed to establish a set of generators specifically designed for this type of dataset. In the case of images, these resulting datasets can be effectively utilized for classification tasks. However, it is important to note that the approach of Guidotti is limited to a single pattern for each image. As a result, the evaluation of explanations is limited to their spatial location. Mamalakis \emph{et al.}~\cite{mamalakis2022investigating} formalized this kind of dataset and proposed another one, which was a collection of binary images of circles and squares with a synthetic classification label indicating whether there were more amounts of pixels from of circles than squares and the contrary. They only verified whether the features have a negative or positive impact on the classification. However, similar to Arras \emph{et al.}~\cite{arras2020ground} and Guidotti~\cite{guidotti_evaluating_2021}, did not consider the value of the explanation.

Aiming to be able to improve the quality of the explanation, we wanted to objectively identify the overall fidelity of 13 different XAI methods, using the methodology proposed by Miró-Nicolau \emph{et al.}~\cite{miro2022evaluating}. We summarize our contributions of this paper as follows:
\begin{itemize}
    \item We define three new datasets with ground truth for the explanations, following our previous proposed methodology~\cite{miro2022evaluating}: $\mli{TXUXIv1}$, $\mli{TXUXIv2}$ and $\mli{TXUXIv3}$.
    \item We objectively compare the fidelity of thirteen well-known XAI methods.
\end{itemize}

The rest of this paper is organized as follows. In the next section, we study the Post-hoc XAI method state-of-art. In Section~\ref{sec:experimental}, we specify the experimental environment, and we describe the datasets, measures, metrics, and predictive models used for experimentation. Section~\ref{sec:results} discuss the results and comparison experiments obtained after applying the two datasets generated using the proposed method to multiple common XAI methods. Finally, in Section~\ref{sec:conclusion} we present the conclusions of the study.

\section{XAI methods}\label{sec:state-of-art}

In this paper, our aim is to compare various XAI methods to determine their fidelity levels. To provide a context for our analysis, we will briefly review the current state-of-the-art for XAI methods in this section. A summary of the whole section can be seen on \cref{tab:state-of-art}.

\begin{table}[htb]
\begin{tabular}{ccccc}
\hline
Article                             & Name                 & Authors                    & Year           & Category                   \\ \hline
\cite{ribeiro2016should}            & LIME                 & Ribeiro \emph{et al.}      & 2016                & Sensit. Analysis       \\
\cite{lundberg_unified_2017}        & SHAP                 & Lundberg \emph{et al.}     & 2017                & Sensit. Analysis       \\
\cite{Petsiuk2018}                  & RISE                 & Petsiuk \emph{et al.}      & 2018                & Sensit. Analysis       \\
\cite{zhou2016learning}             & CAM                  & Zhou \emph{et al.}         & 2016                & CAM                        \\
\cite{selvaraju2017grad}            & GradCAM              & Selvaraju \emph{et al.}    & 2017                & CAM                        \\
\cite{chattopadhay2018grad}         & GradCAM++            & Chattopadhay \emph{et al.} & 2018                & CAM                        \\
\cite{wang2020score}                & ScoreCAM             & Wang \emph{et al.}         & 2020                & \begin{tabular}[c]{@{}c@{}}Sensit. Analysis \\ \& CAM\end{tabular} \\
\cite{muddamsetty2022visual}        & SIDU                 & Muddamsetty \emph{et al.}  & 2022                & \begin{tabular}[c]{@{}c@{}}Sensit. Analysis \\ \& CAM\end{tabular} \\
\cite{simonyan2014very}             & Gradient             & Simonyan \emph{et al.}     & 2014                & Backpropagation            \\
\cite{springenberg2014striving}     & GBP                  & Springerberg \emph{et al.} & 2014                & Backpropagation            \\
\cite{bach2015pixel}                & LRP                  & Bach \emph{et al.}         & 2015                & Backpropagation            \\
\cite{shrikumar2017learning}        & DeepLift             & Shrikumar \emph{et al.}    & 2017                & Backpropagation            \\
\cite{sundararajan2017axiomatic}    & Int. Gradients       & Sundararajan \emph{et al.} & 2017                & Backpropagation            \\
\cite{smilkov2017smoothgrad}        & SmoothGrad           & Smilkov \emph{et al.}      & 2017                & Backpropagation            \\ \hline
\end{tabular}
\caption{Summary of the XAI state-of-art, the methods are categorized according to its internal mechanisms.}
\label{tab:state-of-art}
\end{table}

The initial approaches to obtain explanations are based on analysing the impact of occluding input data on the model's output. This approach is called sensitivity analysis. Ribeiro \emph{et al.}\cite{ribeiro2016should} proposed the Local Interpretable Model Agnostic Explanations (LIME). LIME involves training a transparent model, called a subrogate, to predict the importance of each occluded part. Lundberg \emph{et al.}~\cite{lundberg_unified_2017} proposed a new method, SHapley Additive exPlanations (SHAP), that aimed to unify the previous XAI methods with the usage of the Shapley values from game theory. The authors, aiming to reduce the computational cost of the Shapely values, proposed Kernel SHAP, that used the exact same algorithm and formulation of LIME \cite{ribeiro2016should} introducing the Shapley values into the optimization of the subrogated model. Petsiuk \emph{et al.}\cite{Petsiuk2018} proposed Randomized Input Sampling for Explanation (RISE). Unlike LIME and SHAP, RISE does not use a subrogate model, but instead considers the importance of the difference between the output between the original input and the occluded one directly. 

Another approach found in the literature is the so-called Class Activation Maps (CAM). These methods are based on the original work of Zhou \emph{et al.}~\cite{zhou2016learning}, CAM method consists on adding a global average pooling, a~well-known technique in the field of deep learning, between the feature extractor of a Convolutional Neural Network (CNN) and the classification part. The main drawback of this method is that it modifies the original architecture, and, consequently, modifies the performance of the model. To solve this problem, a set of new methods appeared: GradCAM by Selvaraju~\emph{et al.}\cite{selvaraju2017grad}, that used the gradient of the output with respect to the last convolutional layer to get the importance of each activation map. Chattopadhay~\emph{et al.}\cite{chattopadhay2018grad} proposed GradCAM++, which uses the second derivative instead of the gradient.

Recently, two approaches have been developed that combine the aspects of CAM and occlusion-based methods. Wang et al. \cite{wang2020score} introduced a novel gradient-free CAM method called ScoreCAM. Unlike traditional approaches, ScoreCAM avoids utilizing gradients for generating explanations, due to well-known problems of the gradient calculation as its saturation. Instead, the authors proposed a different approach by observing the changes in the output between the original input and one with only a specific region of interest. The region to perturb is determined by the pixel value from the activation map of the layer that is explained. Muddamsetty \emph{et al.}~\cite{muddamsetty2022visual} proposed the Similarity Difference and Uniqueness (SIDU) method, which occludes the input based on the activation maps of the last convolutional layer and then calculates a weighted sum of these maps using the SIDU metric.

Finally, there are a set of methods aimed to explain the model via back propagating the output results to the input. The initial work of this kind was proposed by Simonyan \emph{et al.}\cite{simonyan2014very}, these authors proposed as the explanation the gradient of the output of the model with respect to the input. Similarly, Springenberg \emph{et al.}\cite{springenberg2014striving} proposed the Guided Backpropagation (GBP) method, this model only differs from the proposal of Simonyan \emph{et al.} on how to handle the non-linearities found on a model, as the ReLU. Bach \emph{et al.}\cite{bach2015pixel} proposed to apply a backwards propagation mechanism sequentially to all layers, with predefined rules for each kind of layer. The rules are based on the conservation of total relevance axiom, which ensures that the sum of relevance assigned to all pixels equals the class score produced by the model. They named this approach Layer Relevance Propgation (LRP). Shrikumar \emph{et al.}~\cite{shrikumar2017learning} introduced Deep Learning Important FeaTures (DeepLift) aiming to improve the rules proposed by Bach \emph{et al.}. The authors incorporated two new axioms based on the existence of a reference baseline value. The reference “input represents some default or ‘neutral’ input that is chosen according to what is appropriate for the problem at hand”. The first axiom, the conservation of total relevance, that ensured that the difference between the relevance assigned to the inputs and the baseline value must equate to the difference between the score of the input image and the baseline value. The second axiom, the chain rule, establishes that relevance must follow the chain rule, akin to gradients. Integrated Gradients (IG) by Sundararajan \emph{et al.}\cite{sundararajan2017axiomatic}, calculates the integral of the space defined between the gradients of the original input and the baseline data, usually a greyscale image. One challenge encountered in gradient-based methods is the generation of noisy results. To mitigate this issue, Smilkov \emph{et al.}~\cite{smilkov2017smoothgrad} proposed an approach known as SmoothGrad. The objective of SmoothGrad is to reduce the noise present in the saliency maps generated using gradient-based methods. This technique involves smoothing the gradient of the model's output with respect to the input by introducing Gaussian noise of varying magnitudes to the input. Subsequently, the saliency maps obtained from each perturbed input are averaged to produce a more refined saliency map.

Once we already reviewed the existing XAI state-of-art, in the following section we defined an experimental setup to compare these methods.

\section{Experimental setup}\label{sec:experimental}

We designed the experimental setup presented in this paper with the aim of comparing thirteen different state-of-the-art eXplainable Artificial Intelligence (XAI) methods and providing insights into their quality. To ensure a fair comparison, we generated three new datasets using the methodology proposed by Mir\'{o}-Nicolau \emph{et al.}~\cite{miro2023novel}. This allowed us to evaluate the performance of the XAI methods under a range of conditions and provide a more comprehensive assessment of their capabilities.

\subsection{Datasets}

In this experimentation, we utilized the methodology we proposed in our previous work~\cite{miro2023novel} that allowed us to generate a data set with explanations ground truth. Subsequently, we employed this methodology to generate three datasets with ground truth for explanations, named TeXture Under eXplainable Insights v1 ($\mli{TXUXIv1}$), $\mli{TXUXIv2}$, and $\mli{TXUXIv3}$. This approach ensured that the datasets were consistent and that the ground truth was accurately defined, enabling us to draw robust conclusions about the performance of the XAI methods under investigation.

\subsection{TeXture Under eXplainable Insights (TXUXI)}

The datasets used in the original paper of  Mir\'{o}-Nicolau \emph{et al.}~\cite{miro2023novel} were characterized by two key features: the availability of ground truth and the simplicity of the images. These features are closely related, as the ground truth is obtained through simplicity. In the original proposal, the problem was simplified to the essentials to test the proposed methodology. However, now that the methodology is validated, more complex datasets can be constructed.

\begin{figure}[h!]
	\centering
	\subfloat[Examples of an image from the $\mli{AIXI-Shape}$ dataset, proposed by Miró-Nicolau \emph{et al.}~\cite{miro2022evaluating}.]{\includegraphics[width=0.45\textwidth]{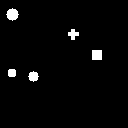}%
	\label{subfigure:aixi-shape}}
	\hfil
	\subfloat[Examples of an image from a $\mli{TXUXI}$ dataset.]{\includegraphics[width=0.45\textwidth]{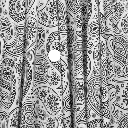}%
	\label{subfigure:txuxi_ex}}
	\caption{Comparison between an image from the proposal of Miró-Nicolau \emph{et al.}~\cite{miro2022evaluating} and the proposal of this paper.}
	\label{figure:txuxi_vs_aixi}
\end{figure}

To this end, we proposed a new family of datasets named TeXture Under eXplainable Insights ($\mli{TXUXI}$), based on the AIXI dataset proposed by Mir\'{o}-Nicolau \emph{et al.}~\cite{miro2022evaluating}. These datasets are created using a texture as the background, which introduces additional complexity and challenges for the XAI methods. By using textures, we can evaluate the performance of the XAI methods under more realistic and diverse conditions, providing a more comprehensive assessment of their capabilities. Figure \ref{figure:txuxi_vs_aixi} allowed us to see the difference between the original proposal of Mir\'{o}-Nicolau \emph{et al.}~\cite{miro2022evaluating} and the proposed dataset of this work. 

\begin{figure}[h!]
	\centering
	\subfloat[Examples of an image from the $\mli{TXUXIv1}$ dataset. All images have exactly the same background.]{\includegraphics[width=0.3\textwidth]{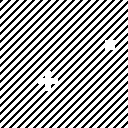}%
	\label{subfigure:txuxiv1}}
	\hfil
	\subfloat[Examples of an image from the $\mli{TXUXIv2}$ dataset. All images have exactly the same background.]{\includegraphics[width=0.3\textwidth]{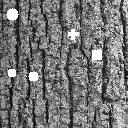}%
	\label{subfigure:txuxiv2}}
	\hfil
	\subfloat[Examples of an image from the $\mli{TXUXIv3}$ dataset. The background for each image is selected from a pool of 5640 available textures from the Describable Textures Dataset (DTD)~\cite{cimpoi14describing}.]{\includegraphics[width=0.3\textwidth]{figs/v3.png}%
	\label{subfigure:txuxiv3}}
	\caption{Examples from all the $\mli{TXUXI}$ datasets.}
	\label{figure:txutxi}
\end{figure}

We defined three different datasets of this family. The element that differentiates them is the texture used as the background, with each one with the same shapes from the original AIXI-Shape dataset, proposed by  Mir\'{o}-Nicolau \emph{et al.}~\cite{miro2023novel}, but with a background with an increased level of complexity:

\begin{enumerate}
\item \textbf{TXUXI Version 1 ($\mli{TXUXIv1}$)}: This dataset features a background composed of perfect lines, where each line has a binary value of either 1 or 0. The lines are arranged in a regular pattern, creating a simple yet structured texture. An example of an image from the ($\mli{TXUXIv1}$) dataset is shown in \Cref{subfigure:txuxiv1}. This dataset serves as the baseline for the more complex textures used in the TXUXI family.
\item \textbf{TXUXI Version 2 ($\mli{TXUXIv2}$)}: This dataset features a more complex background, specifically a wood tree texture extracted from the Describable Textures Dataset (DTD)~\cite{cimpoi14describing}. All images in the dataset have the exact same background. An example of an image from the $\mli{TXUXIv2}$ dataset is shown in \Cref{subfigure:txuxiv2}.
\item \textbf{TXUXI Version 3 ($\mli{TXUXIv3}$)}: This dataset used similarly to $\mli{TXUXIv2}$ as background of the images textures from the DTD dataset. However, in this case all textures are used, with a total of 5640 different possible backgrounds. An example of an image from the $\mli{TXUXIv3}$ dataset is shown in \Cref{subfigure:txuxiv3}.
\end{enumerate}

The textures from the $\mli{TXUXIv2}$ and $\mli{TXUXIv3}$ are obtained from the Describable Textures Dataset (DTD)~\cite{cimpoi14describing}. This dataset is a collection of, 5640 different textures. 

All three datasets, in addition to their generation scripts, can be found in \url{https://github.com/miquelmn/aixi-dataset/releases/tag/1.5.0}.

\subsection{Methods} \label{sec:methods}

In the previous sections, we explored the current state-of-the-art XAI methods, we compared all of them, in particular these thirteen methods are: LIME~\cite{ribeiro2016should}, Kernel SHAP~\cite{lundberg_unified_2017}, RISE~\cite{Petsiuk2018},  GradCAM~\cite{selvaraju2017grad}, GradCAM++~\cite{chattopadhay2018grad}, ScoreCAM~\cite{wang2020score}, SIDU~\cite{muddamsetty2022visual}, Gradient~\cite{simonyan2014very}, GBP~\cite{springenberg2014striving}, LRP~\cite{bach2015pixel}, DeepLift~\cite{shrikumar2017learning}, Integrated Gradients~\cite{sundararajan2017axiomatic} and SmoothGrad~\cite{smilkov2017smoothgrad}. We have only discarded the original CAM~\cite{zhou2016learning} method due to the need to add a new layer to the original model, retrain it, and modify the overall AI model performance to explain.

We used publicly available implementations provided by the authors or the Captum~\cite{kokhlikyan2020captum} library. We have only implemented one of the methods, SIDU, the implementation is available at \url{https://github.com/miquelmn/sidu_torch}. We used the default configurations and hyperparameters for all methods.

\subsection{AI model}

The methods we wanted to compare are post-hoc XAI methods, for this reason an artificial intelligence model is needed from to extract the explanations. We used a Convolutional Neural Network (CNN) as our model. 

The basis of the modern CNN were introduced by Krizhevsky \emph{et al.}~\cite{krizhevsky2012imagenet}. The authors have proposed a model comprising two distinct components: a feature extraction part responsible for recognizing patterns, and a classification part that consolidates the recognized patterns into meaningful semantic information. The feature extraction part comprises three key elements: convolutional layers, rectified linear unit (ReLU) activations, and max pooling operations. On the other hand, the classification component is implemented as a multilayer perceptron (MLP) with a softmax activation function. The original work provides a detailed discussion on each of these building blocks. This model was designed to solve image related tasks, and it was designed to be applicable for both classification and regression tasks.

The specific details of the implementation are beyond the scope of this study. However, the weights and the architecture used are available at \url{https://github.com/miquelmn/aixi-dataset/releases/tag/1.5.0}.

\subsection{Performance measures} \label{sec:measures}

To accomplish our objective of conducting a fair comparison of various XAI methods, selecting appropriate performance measures to evaluate and compare the methods was crucial. In this study, we utilized multiple measures that allowed us to evaluate the similarity between the predicted saliency map and its corresponding ground truth. This issue has been thoroughly explored in the literature, particularly in the review conducted by Riche \emph{et al.}~\cite{riche2013saliency}. Judd \emph{et al.}~\cite{judd2012benchmark} also analyse a set of metrics to compare different saliency maps. 

Riche \emph{et al.}~\cite{riche2013saliency} analysed the current literature on saliency map comparison and found that treating both saliency maps as probability distributions was a critical factor in the comparison process, this enabled us to employ well-established measures. We utilized two distinct measures, which were originally reviewed by Riche \emph{et al.}~\cite{riche2013saliency} or used by Judd \emph{et al.}~\cite{judd2012benchmark}:

\begin{itemize}
    \item Earth Mover Distance (EMD). This measure the distance between two probability distributions. EMD can be defined as the minimum amount of work required to transform one distribution into the other. See \cref{eq:emd}.
    \item Similarity Metric (MIN). First introduced by \cite{judd2012benchmark}, indicated how similar two probability distributions defined as the sum of the minimum values at each point in both distributions. See \cref{eq:min} for more details. 
\end{itemize}

The previous measure must be analysed differently, bearing in mind that while EMD is a distance, with the perfect value represented by a 0 value, the MIN is a metric, meaning that the perfect value represent by a value of 1,

\begin{equation}
    \begin{aligned}
        \mli{EMD} = {} & (\min_{f_{ij}} \sum_{i,j} f_{ij} \cdot d_{ij}) + | \sum_{i} \mli{SMGT}_{i} - \sum_{j} SM_{j}| \max_{i,j} d_{i,j} \\
                & s.t. f_{i,j} \geq 0, \sum_{j} f_{i,j} \leq \mli{SMGT}_{i}, \sum_{i} f_{i,j} \leq \mli{SM}_{j} \\
                & \sum_{i, j} f_{i, j} = \min(\sum_{i} \mli{SMGT}_{i} - \sum_{j} \mli{SM}_{j})
    \end{aligned}
    \label{eq:emd}
\end{equation}

\begin{equation}
   \mli{MIN} = \sum_{x=1}^{X} \min (\mli{SM}(x), \mli{SMGT}(x))
    \label{eq:min}
\end{equation}

\noindent where each $f_{ij}$ represents the amount transported from the $i_{th}$ supply to the $j_{th}$ demand. $d_{ij}$ is the ground distance between the sample $i$ and sample $j$ in the distribution. $\mli{SMGT}$ is the probability distribution of the ground truths and $\mli{SM}$ the probability distribution of the predicted saliency map. 

\subsection{Experiments}\label{sec:experiments}

We conducted three experiments to compare the XAI methods selected in \cref{sec:methods}, each one defined by a different dataset. The first experiment used the $\mli{TXUXIv1}$ dataset, this experiment worked as a baseline to evaluate the performance of the other experiments and results. The second experiment used the more complex $\mli{TXUXIv2}$ dataset, and the third used the even more complex $\mli{TXUXIv3}$ dataset.

Each experiment aimed to provide a fair comparison of the XAI methods. We used all metrics showed in \cref{sec:measures} in each experiment and a CNN. We independently trained the same model for each dataset, and achieved nearly perfect prediction results for each one of them. This is an essential prerequisite for being able to fairly compare different methods. 

\section{Results and discussion}\label{sec:results}
In this section, we discuss and analyse the results of the experiments defined in section \ref{sec:experiments}. 

\subsection{Experiment 1: $\mli{TXUXIv1}$}

\begin{figure}[!ht]
	\centering
	\subfloat[]{\includegraphics[width=0.85\textwidth]{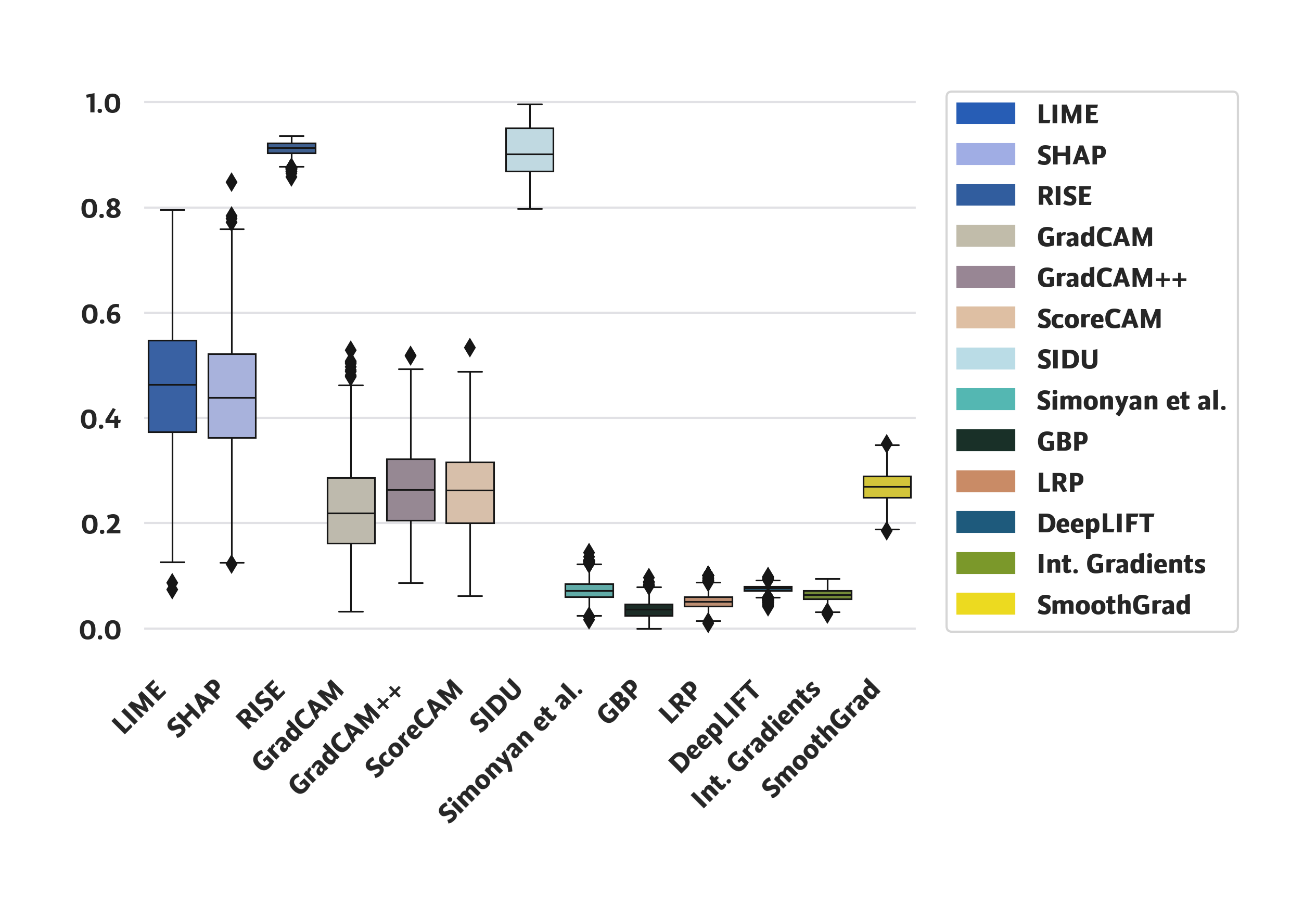}%
	\label{subfigure:v1_bp_emd}}
	\hfil
	\subfloat[]{\includegraphics[width=0.85\textwidth]{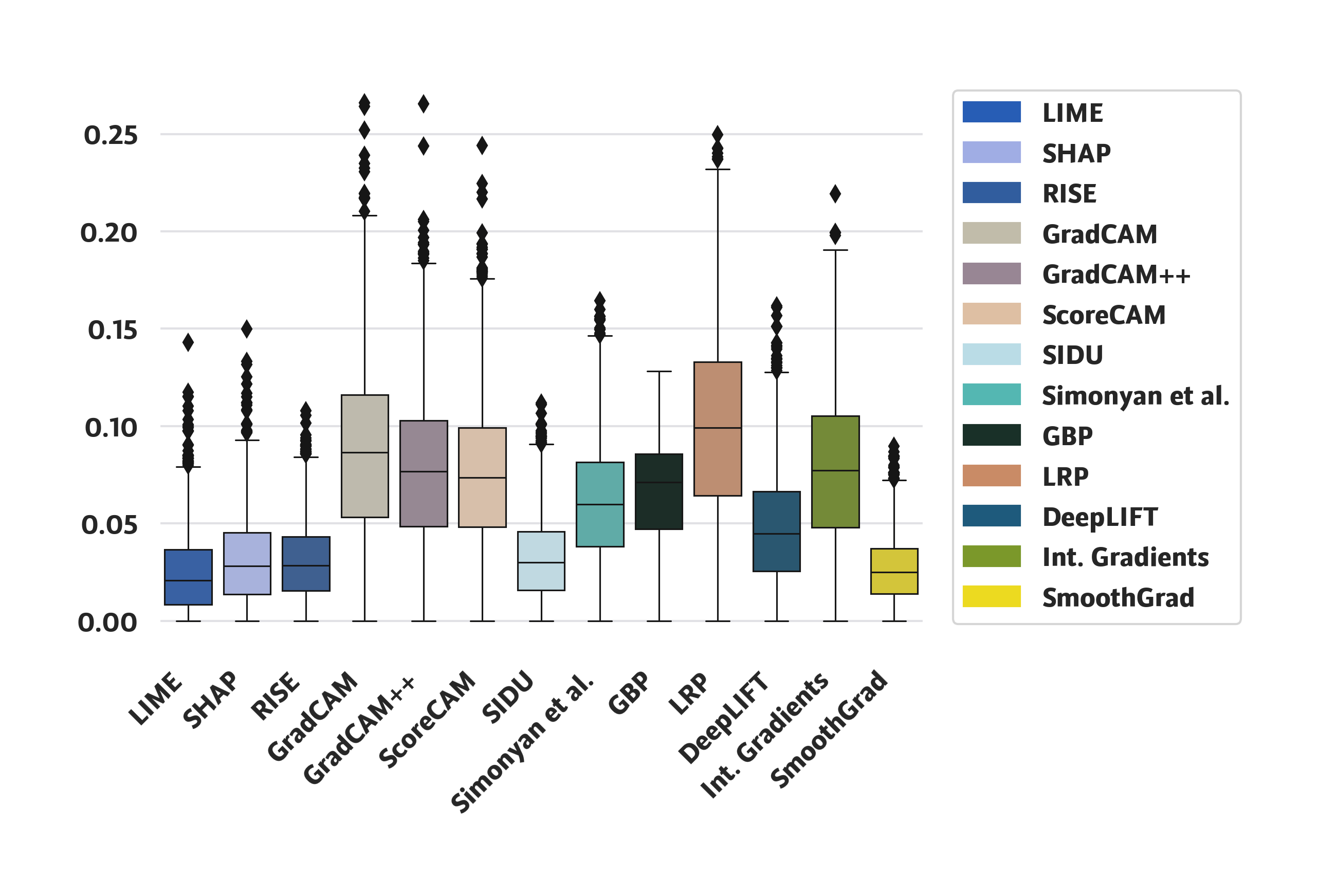}}%
	\label{subfigure:v1_bp_min}
	\caption{Box-plot for the $\mli{EMD}$ (a), and $\mli{MIN}$ (b) metrics obtained in the first experiment.}
	\label{figure:v1_bp}
\end{figure}

Table \ref{tab:results_first_experiment} and Fig. \ref{figure:v1_bp} list the results obtained in the first experiment. The table shows the results for both metrics, $\mli{EMD}$ and $\mli{MIN}$, and highlights the best-performing method among the thirteen methods compared methods. Both metrics were calculated image-wise, we aggregated the result with the mean and the standard deviation. While the figure is a box-plot representing the same data visually.

\begin{table}[!htb]
\centering
\begin{tabular}{lcccc}
\hline  
 Method                                          & Ranking   & $\mli{EMD}$                 & Ranking       & $\mli{MIN}$                \\
\hline
 LIME~\cite{ribeiro2016should}                   & 11 & $0.460 \pm 0.116$ & 12 & $0.025 \pm 0.02$  \\
 SHAP~\cite{lundberg_unified_2017}               & 10 & $0.442 \pm 0.117$ & 11 & $0.031 \pm 0.023$ \\
 RISE~\cite{Petsiuk2018}                         & 13 & $0.913 \pm 0.014$ & 10 & $0.031 \pm 0.02$  \\
 GradCAM~\cite{selvaraju2017grad}                & 6 & $0.225  \pm 0.086$ & 2 & $0.084 \pm 0.048$ \\
 GradCAM++~\cite{chattopadhay2018grad}           & 8 & $0.263  \pm 0.081$ & 4 & $0.076 \pm 0.041$ \\
 ScoreCAM~\cite{wang2020score}                   & 7 & $0.257  \pm 0.087$ & 5 & $0.075 \pm 0.039$ \\
 SIDU~\cite{muddamsetty2022visual}               & 12 & $0.908 \pm 0.049$ & 9 & $0.032 \pm 0.021$ \\
 Simonyan \emph{et al.}~\cite{simonyan2014very}  & 4 & $0.074  \pm 0.021$ & 6 & $0.061 \pm 0.032$ \\
 GBP~\cite{springenberg2014striving}             & \textbf{1} & $\mathbf{0.035 \pm 0.017}$ & 7 & $0.06 \pm 0.035$  \\
 LRP~\cite{bach2015pixel}                        & 2 & $0.053  \pm 0.016$ & \textbf{1} & $\mathbf{0.099 \pm 0.051}$ \\
 DeepLIFT~\cite{shrikumar2017learning}           & 5 & $0.075  \pm 0.008$ & 8 & $0.048 \pm 0.03$  \\
 Int. Gradients~\cite{sundararajan2017axiomatic} & 3 & $0.064  \pm 0.012$ & 3 & $0.077 \pm 0.041$ \\
 SmoothGrad~\cite{smilkov2017smoothgrad}         & 9 & $0.270  \pm 0.029$ & 13 & $0.027 \pm 0.017$ \\
\hline                                         
\end{tabular}
\caption{Mean and standard deviation obtained in the first experiment with the $\mli{TXUXIv1}$ dataset for both $\mli{EMD}$ and $\mli{MIN}$ metrics. The ranking columns indicate the order of methods according to the respective metric mean.}
\label{tab:results_first_experiment}
\end{table}

\begin{figure}[!htb]
	\centering
     	\subfloat[Imagre from the $\mli{TXUXIv1}$ dataset]{\includegraphics[width=0.2\textwidth]{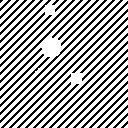}%
    	\label{subfigure:image}}
        \hfil
        \subfloat[LIME~\cite{ribeiro2016should} result.]{\includegraphics[width=0.2\textwidth]{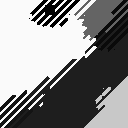}%
    	\label{subfigure:lime}}
        \hfil
        \subfloat[SHAP~\cite{lundberg_unified_2017} result.]{\includegraphics[width=0.2\textwidth]{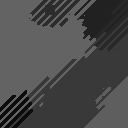}%
    	\label{subfigure:lime}}
        \hfil
        \subfloat[RISE~\cite{Petsiuk2018} result.]{\includegraphics[width=0.2\textwidth]{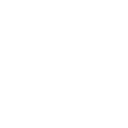}%
    	\label{subfigure:rise}}
    	\\
    	\subfloat[GradCAM~\cite{selvaraju2017grad} result.]{\includegraphics[width=0.2\textwidth]{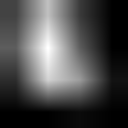}%
    	\label{subfigure:grad_cam}}
    	\hfil
    	\subfloat[GradCAM++\cite{chattopadhay2018grad} result.]{\includegraphics[width=0.2\textwidth]{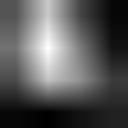}%
    	\label{subfigure:grad_cam_plus}}
        \hfil
        \subfloat[ScoreCAM~\cite{wang2020score} result.]{\includegraphics[width=0.2\textwidth]{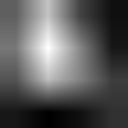}%
    	\label{subfigure:score_cam}}
    	\hfil
        \subfloat[SIDU~\cite{muddamsetty2022visual} result.]{\includegraphics[width=0.2\textwidth]{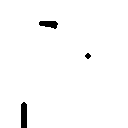}%
    	\label{subfigure:sidu}}
        \\
    	\subfloat[Simonyan~\cite{simonyan2014very} result.]{\includegraphics[width=0.2\textwidth]{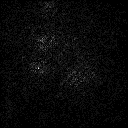}%
    	\label{subfigure:gradient}}
        \hfil
        \subfloat[GBP~\cite{springenberg2014striving} result.]{\includegraphics[width=0.2\textwidth]{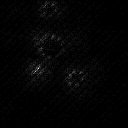}%
    	\label{subfigure:gbp}}
    	\hfil
        \subfloat[LRP~\cite{bach2015pixel} result.]{\includegraphics[width=0.2\textwidth]{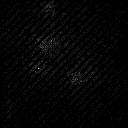}%
    	\label{subfigure:lrp}}
    	\hfil
        \subfloat[SmoothGrad~\cite{smilkov2017smoothgrad} result.]{\includegraphics[width=0.2\textwidth]{figs/v1_cmp/lrp.png}%
    	\label{subfigure:smooth}}
    	\\
        \subfloat[DeepLIFT~\cite{shrikumar2017learning} result.]{\includegraphics[width=0.2\textwidth]{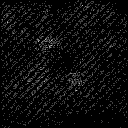}%
    	\label{subfigure:deep_lift}}
        \hfil
    	\subfloat[Integrated Gradients~\cite{sundararajan2017axiomatic} result.]{\includegraphics[width=0.2\textwidth]{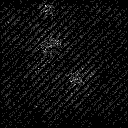}%
    	\label{subfigure:ig}}
	\caption{Results to apply different XAI methods to a $\mli{TXUXIv1}$ sample in subfigure (a).}
	\label{figure:cmp_v1}
\end{figure}

From Table \ref{tab:results_first_experiment} and Figs. \ref{figure:cmp_v1}, \ref{figure:v1_bp}  we can see that the LRP, proposed by Bach \emph{et al.}~\cite{birhane2021large}, stands out for its good results in both metrics, being the best method according to $\mli{MIN}$ and second best according to $\mli{EMD}$. However, the rest of methods performance largely differ between both metrics. This behaviour is caused to the higher sensitivity of the $\mli{MIN}$ metric to noise: small activated values in the expected empty zone. On the other hand, $\mli{EMD}$ ignores this noise and give more importance to the highly activated inputs. In particular, we can see that the backpropagation methods, apart from LRP~\cite{birhane2021large}, are the best methods to generate correct saliencies, although they tend to generate noise. This is accentuated due to the background of the $\mli{TXUXIv1}$ dataset images, it causes large amounts of noise because it has the maximum activation value. We can also see that all the good methods had small dispersions, indicated by the standard deviation, while some methods with bad results also had large dispersions, as LIME~\cite{ribeiro2016should} and SHAP~\cite{lundberg_unified_2017}. It is also interesting to mention that in this scenario even SmoothGrad~\cite{smilkov2017smoothgrad} that specifically aims to remove this noise, generated it. In conclusion, it is clear that only a method that obtained good results in both metrics is a truly good method with high fidelity and low noise.

In \cref{figure:cmp_v1}, we can observe examples of the results obtained for all methods on an image from the $\mli{TXUXIv1}$ dataset. The explanations obtained are consistent with the metric results. As expected, we can see that the methods that rely on some degree to sensitivity analysis (RISE~\cite{Petsiuk2018}, LIME~\cite{ribeiro2016should}, SIDU~\cite{muddamsetty2022visual}, ScoreCAM~\cite{wang2020score}) perform poorly due to the ease of generating out-of-distribution (OOD) samples with this dataset and the need for extensive fine-turning on the occlusion nature (value, area, etc.). On the other hand, we can also see that the methods based on back propagating the output to the input (Simonyan \emph{et al.}~\cite{simonyan2014very}, LRP~\cite{bach2015pixel}, GBP~\cite{springenberg2014striving}, and Integrated Gradient~\cite{sundararajan2017axiomatic}) have the best results on the $\mli{EMD}$ metric, indicating that they better capture the explanation. According to the $\mli{MIN}$ metric these methods also have the best results, however are very low values, provoked for the tendency to produce more noise. GradCAM~\cite{selvaraju2017grad} and GradCAM++~\cite{chattopadhay2018grad} are comparatively much worse than the backpropagating methods, and at the same time, they are significantly better than the occlusion methods.

\subsection{Experiment 2: $\mli{TXUXIv2}$}

\begin{figure}[!ht]
	\centering
	\subfloat[]{\includegraphics[width=0.85\textwidth]{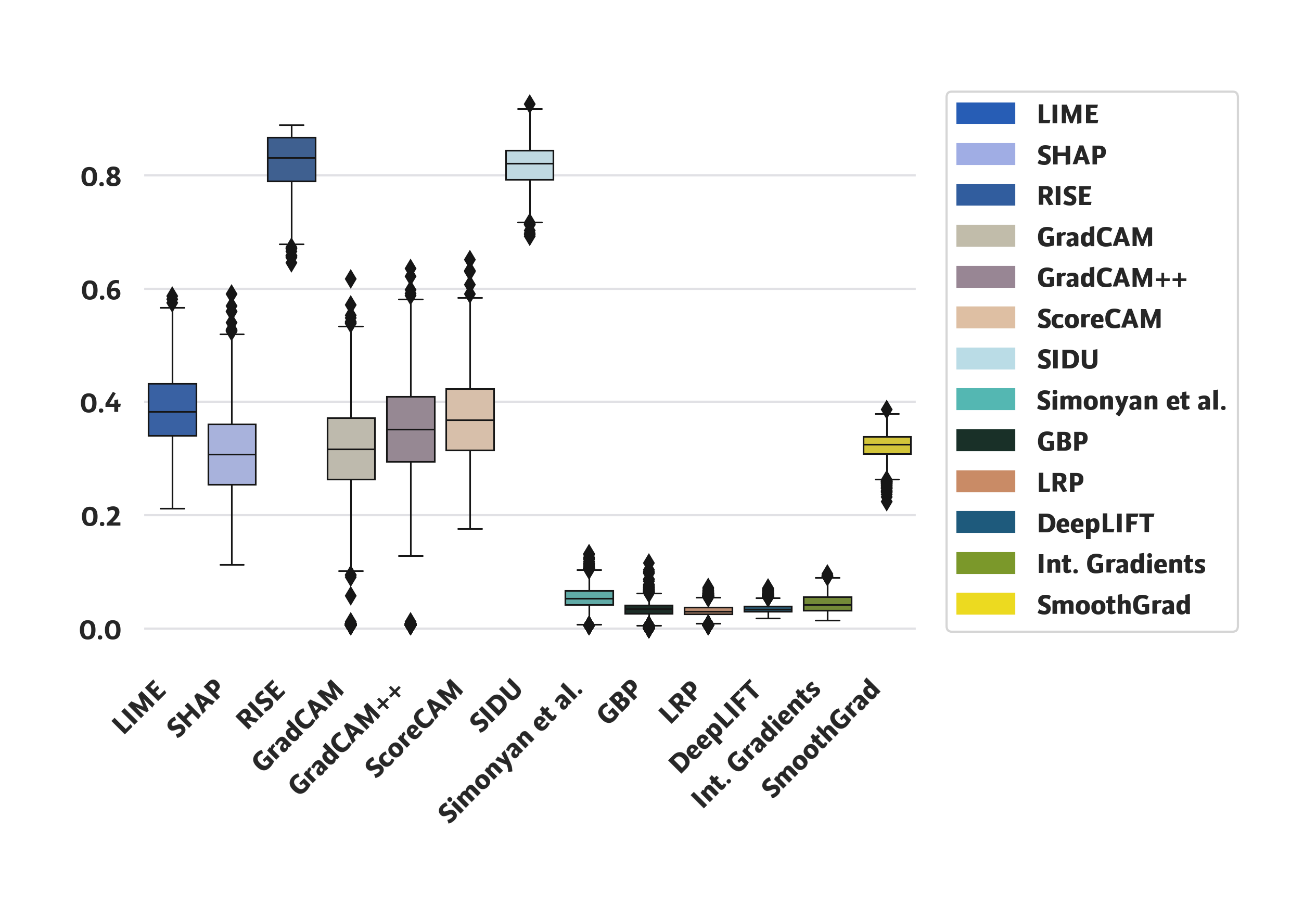}%
	\label{subfigure:v2_bp_emd}}
	\hfil
	\subfloat[]{\includegraphics[width=0.85\textwidth]{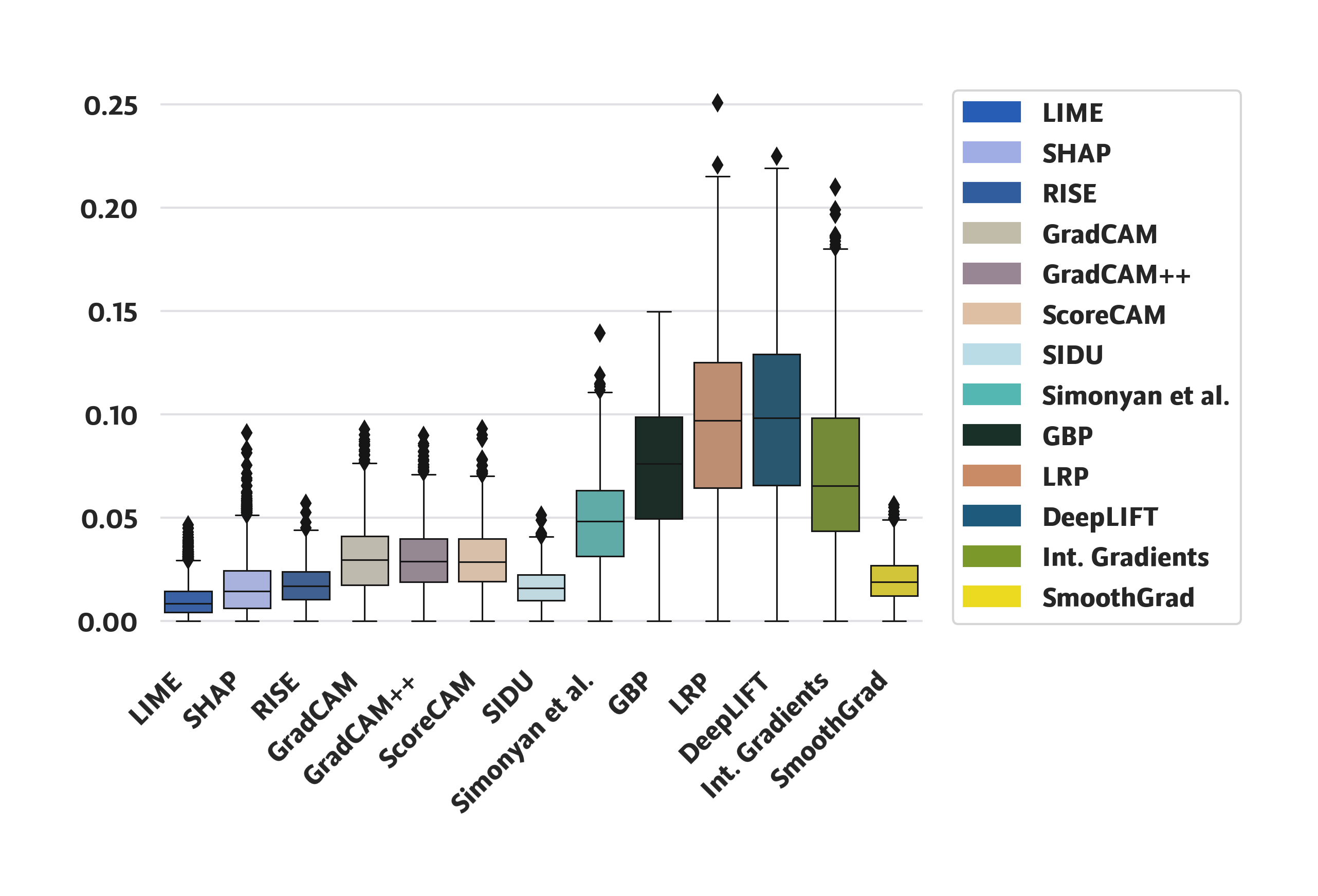}}%
	\label{subfigure:v2_bp_min}
	\caption{Box-plot for the $\mli{EMD}$ (a), and $\mli{MIN}$ (b) metrics obtained in the second experiment.}
	\label{figure:v2_bp}
\end{figure}

Table \ref{tab:results_second_experiment} and \cref{figure:v2_bp} list the results obtained in the second experiment. The table show, once again, the results of both metrics, $\mli{EMD}$ and $\mli{MIN}$, and highlights which of the compared methods obtained the best results. While the figure is a box-plot representing the same data visually.

In contrast to the previous experiment, we did not expect the presence of any gap between both metrics. The fact that the images backgrounds in this experiment do not have the maximum possible activation value alleviates this problem. We can see that the bests methods, are once again the backpropagation-based ones, except for SmoothGrad~\cite{smilkov2017smoothgrad}, followed by the rest of methods with a significant gap. We can also see that backpropagation-based methods, again, have small dispersion of the metrics (standard deviation), while the rest of methods had higher dispersion values. 

\begin{table}[!htb]
\centering
\begin{tabular}{lcccc}
\hline  
 Method                                          & Ranking   & $\mli{EMD}$          & Ranking   & $\mli{MIN}$       \\
\hline
 LIME~\cite{ribeiro2016should}                   & 11       & $0.387 \pm 0.066$     & 13        & $0.010 \pm 0.008$     \\
 SHAP~\cite{lundberg_unified_2017}               & 6        & $0.310 \pm 0.076$     & 12        & $0.017 \pm 0.013$ \\
 RISE~\cite{Petsiuk2018}                         & 13       & $0.823 \pm 0.050$     & 9         & $0.017 \pm 0.010$     \\
 GradCAM~\cite{selvaraju2017grad}                & 7        & $0.313 \pm 0.093$     & 7         & $0.029 \pm 0.018$     \\
 GradCAM++~\cite{chattopadhay2018grad}           & 9        & $0.346 \pm 0.099$     & 8         & $0.029 \pm 0.016$     \\
 ScoreCAM~\cite{wang2020score}                   & 10       & $0.370 \pm 0.077$     & 6         & $0.030 \pm 0.016$     \\
 SIDU~\cite{muddamsetty2022visual}               & 12       & $0.819 \pm 0.038$     & 11        & $0.016 \pm 0.009$     \\
 Simonyan \emph{et al.}~\cite{simonyan2014very}  & 5        & $0.056 \pm 0.022$     & 5         & $0.047 \pm 0.024$     \\
 GBP~\cite{springenberg2014striving}             & \textbf{1}        & $\mathbf{0.032 \pm 0.016}$     & 4     & $0.068 \pm 0.040$  \\
 LRP~\cite{bach2015pixel}                        & \textbf{1}        & $\mathbf{0.032 \pm 0.011}$     & 2     & $0.094 \pm 0.047$ \\
 DeepLIFT~\cite{shrikumar2017learning}           & 3        & $0.036 \pm 0.009$     & \textbf{1}         & $\mathbf{0.095 \pm 0.046}$ \\
 Int. Gradients~\cite{sundararajan2017axiomatic} & 4        & $0.045 \pm 0.018$     & 3         & $0.073 \pm 0.041$ \\
 SmoothGrad~\cite{smilkov2017smoothgrad}         & 8        & $0.322 \pm 0.024$     & 4         & $0.020 \pm 0.011$ \\
\hline                                         
\end{tabular}
\caption{Mean and standard deviation obtained in the second experiment with the $\mli{TXUXIv2}$ dataset for both $\mli{EMD}$ and $\mli{MIN}$ metrics. The ranking columns indicate the order of methods according to the respective metric mean. }
\label{tab:results_second_experiment}
\end{table}

\begin{figure}[!htb]
	\centering
     	\subfloat[Imagre from the $\mli{TXUXIv2}$ dataset]{\includegraphics[width=0.2\textwidth]{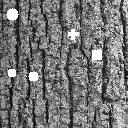}%
    	\label{subfigure:image_v2}}
    	\hfil
        \subfloat[LIME~\cite{ribeiro2016should} results.]{\includegraphics[width=0.2\textwidth]{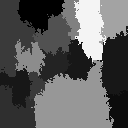}%
    	\label{subfigure:lime_v2}}
        \hfil
        \subfloat[SHAP~\cite{lundberg_unified_2017} results.]{\includegraphics[width=0.2\textwidth]{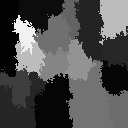}%
    	\label{subfigure:rise_v2}}
        \hfil
        \subfloat[RISE~\cite{Petsiuk2018} results.]{\includegraphics[width=0.2\textwidth]{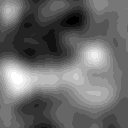}%
    	\label{subfigure:rise_v2}}
    	\\
    	\subfloat[GradCAM~\cite{selvaraju2017grad} results.]{\includegraphics[width=0.2\textwidth]{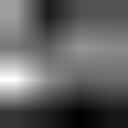}%
    	\label{subfigure:grad_cam_v2}}
    	\hfil
    	\subfloat[GradCAM++\cite{chattopadhay2018grad} results.]{\includegraphics[width=0.2\textwidth]{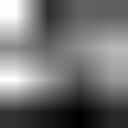}%
    	\label{subfigure:grad_cam_plus_v2}}
        \hfil
        \subfloat[ScoreCAM~\cite{wang2020score} results.]{\includegraphics[width=0.2\textwidth]{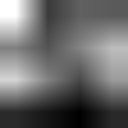}%
    	\label{subfigure:score_cam_v2}}
        \hfil
    	\subfloat[SIDU~\cite{muddamsetty2022visual} results.]{\includegraphics[width=0.2\textwidth]{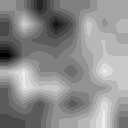}%
    	\label{subfigure:sidu_v2}}
        \\
    	\subfloat[Simonyan~\cite{simonyan2014very} results.]{\includegraphics[width=0.2\textwidth]{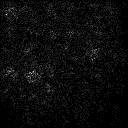}%
    	\label{subfigure:gradient_v2}}
        \hfil
    	\subfloat[GBP~\cite{springenberg2014striving} results.]{\includegraphics[width=0.2\textwidth]{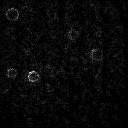}%
    	\label{subfigure:gbp_v2}}
        \hfil
        \subfloat[LRP~\cite{bach2015pixel} results.]{\includegraphics[width=0.2\textwidth]{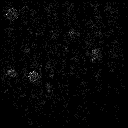}%
    	\label{subfigure:lrp_v2}}
    	\hfil
        \subfloat[DeepLIFT~\cite{shrikumar2017learning} results.]{\includegraphics[width=0.2\textwidth]{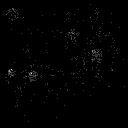}%
    	\label{subfigure:deep_lift_v2}}
        \\
    	\subfloat[Integrated Gradients~\cite{sundararajan2017axiomatic} results.]{\includegraphics[width=0.2\textwidth]{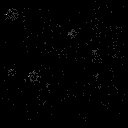}%
    	\label{subfigure:ig_v2}}
        \hfil
        \subfloat[SmoothGrad~\cite{smilkov2017smoothgrad} results.]{\includegraphics[width=0.2\textwidth]{figs/v2_cmp/lrp.png}%
    	\label{subfigure:smooth_v2}}
	\caption{Results to apply different XAI methods to the $\mli{TXUXIv2}$ sample found in subfigure (a).}
	\label{figure:cmp_v2}
\end{figure}

In Figure \ref{figure:cmp_v2} we can see an example of the results obtained for the second experiment. The results are consistent with the metrics, with the back propagation-based methods producing the best results, while the sensitivity analysis based methods (RISE~\cite{Petsiuk2018}, LIME~\cite{ribeiro2016should}, SHAP~\cite{lundberg_unified_2017}, SIDU~\cite{muddamsetty2022visual}, and ScoreCAM~\cite{wang2020score}) the worst ones. We can also see the bad results of SmoothGrad~\cite{smilkov2017smoothgrad}. We consider that the bad results of this method if due to the generation of out of domain inputs with the addition of Gaussian noise. 

The results obtained from this experiment are compatible and coherent with the ones obtained in the previous experiment. On one hand, the best results were obtained by the back-propagation methods, with the caveat of generating noise. On the other hand, the worst methods are the ones based on occlusion, which are very sensitive to OOD samples and the selection of the occlusion method. This sensitivity is exacerbated due to the fact of the few classes of the prediction model. 

\subsection{Experiment 3: $\mli{TXUXIv3}$}

\begin{figure}[!htb]
	\centering
     	\subfloat[Imagre from the $\mli{TXUXIv3}$ dataset]{\includegraphics[width=0.2\textwidth]{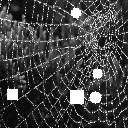}%
    	\label{subfigure:image_v3}}
    	\hfil
        \subfloat[LIME~\cite{ribeiro2016should} results.]{\includegraphics[width=0.2\textwidth]{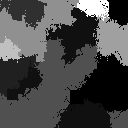}%
    	\label{subfigure:lime_v3}}
        \hfil
        \subfloat[SHAP~\cite{lundberg_unified_2017} results.]{\includegraphics[width=0.2\textwidth]{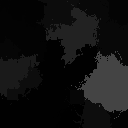}%
    	\label{subfigure:lime_v3}}
        \hfil
        \subfloat[RISE~\cite{Petsiuk2018} results.]{\includegraphics[width=0.2\textwidth]{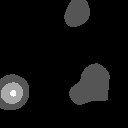}%
    	\label{subfigure:rise_v3}}
    	\\
    	\subfloat[GradCAM~\cite{selvaraju2017grad} results.]{\includegraphics[width=0.2\textwidth]{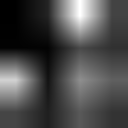}%
    	\label{subfigure:grad_cam_v3}}
    	\hfil
    	\subfloat[GradCAM++\cite{chattopadhay2018grad} results.]{\includegraphics[width=0.2\textwidth]{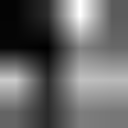}%
    	\label{subfigure:grad_cam_plus_v3}}
        \hfil
        \subfloat[ScoreCAM~\cite{wang2020score} results.]{\includegraphics[width=0.2\textwidth]{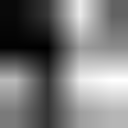}%
    	\label{subfigure:score_cam_v3}}
        \hfil
    	\subfloat[SIDU~\cite{muddamsetty2022visual} results.]{\includegraphics[width=0.2\textwidth]{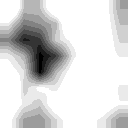}%
    	\label{subfigure:sidu_v3}}
        \\
    	\subfloat[Simonyan~\cite{simonyan2014very} results.]{\includegraphics[width=0.2\textwidth]{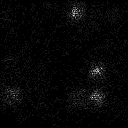}%
    	\label{subfigure:gradient_v3}}
    	\hfil
    	\subfloat[GBP~\cite{springenberg2014striving} results.]{\includegraphics[width=0.2\textwidth]{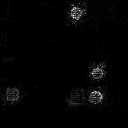}%
    	\label{subfigure:gbp_v3}}
        \hfil
        \subfloat[LRP~\cite{bach2015pixel} results.]{\includegraphics[width=0.2\textwidth]{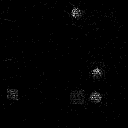}%
    	\label{subfigure:lrp_v3}}
    	\hfil
        \subfloat[DeepLIFT~\cite{shrikumar2017learning} results.]{\includegraphics[width=0.2\textwidth]{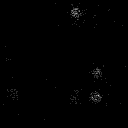}%
    	\label{subfigure:deep_lift_v3}}
        \\
    	\subfloat[Integrated Gradients~\cite{sundararajan2017axiomatic} results.]{\includegraphics[width=0.2\textwidth]{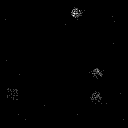}%
    	\label{subfigure:ig_v3}}
        \hfil
        \subfloat[SmoothGrad~\cite{smilkov2017smoothgrad} results.]{\includegraphics[width=0.2\textwidth]{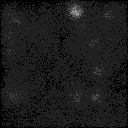}%
    	\label{subfigure:smooth_v3}}

	\caption{Results to apply different XAI methods to the $\mli{TXUXIv3}$ sample found in subfigure (a).}
	\label{figure:cmp_v3}
\end{figure}

The results of the third experiment are shown in Table \ref{tab:results_third_experiment} and Figure \ref{figure:cmp_v3}. The table show, once again, the results of both metrics, $\mli{EMD}$ and $\mli{MIN}$, and highlights which of the thirteen compared methods obtained the best results. While the figure is a box-plot representing the same data visually. 

\begin{table}[!htb]
\centering
\begin{tabular}{lcccc}
\hline  
 Method                                                 & Ranking       & $\mli{EMD}$       & Ranking   & $\mli{MIN}$         \\
\hline
LIME~\cite{ribeiro2016should}                           & 12            & $0.394 \pm 0.093$ & 13 & $0.015 \pm 0.012$ \\
SHAP~\cite{lundberg_unified_2017}                       & 8             & $0.281 \pm 0.113$ & 6 & $0.041 \pm 0.028$ \\
RISE~\cite{Petsiuk2018}                                 & 9             & $0.302 \pm 0.151$ & 7 & $0.039 \pm 0.033$ \\
GradCAM~\cite{selvaraju2017grad}                        & 7             & $0.256 \pm 0.083$ & 8 & $0.038 \pm 0.019$ \\
GradCAM++~\cite{chattopadhay2018grad}                   & 10            & $0.306 \pm 0.086$ & 9 & $0.035 \pm 0.018$ \\
ScoreCAM~\cite{wang2020score}                           & 11            & $0.339 \pm 0.087$ & 10 & $0.034 \pm 0.017$ \\
SIDU~\cite{muddamsetty2022visual}                       & 13            & $0.719 \pm 0.067$ & 12 & $0.029 \pm 0.011$  \\
Simonyan \emph{et al.}~\cite{simonyan2014very}          & 4             & $0.030 \pm 0.016$ & 5 & $0.119 \pm 0.056$ \\
GBP~\cite{springenberg2014striving}                     & \textbf{1}    & $\mathbf{0.019 \pm 0.010}$ & \textbf{1} & $\mathbf{0.245 \pm 0.108}$ \\
LRP~\cite{bach2015pixel}                                & 2             & $0.022 \pm 0.011$ & 2 & $0.202 \pm 0.102$ \\
DeepLIFT~\cite{shrikumar2017learning}                   & 5             & $0.032 \pm 0.018$ & 4 & $0.143 \pm 0.092$ \\
Int. Gradients~\cite{sundararajan2017axiomatic}         & 3             & $0.030 \pm 0.017$ & 3 & $0.144 \pm 0.085$ \\
SmoothGrad~\cite{smilkov2017smoothgrad}                 & 6             & $0.249 \pm 0.050$ & 11 & $0.025 \pm 0.016$ \\
\hline                                         
\end{tabular}
\caption{Mean and standard deviation (value after $\pm$) obtained in the third experiment with the $\mli{TXUXIv3}$ dataset for both $\mli{EMD}$ and $\mli{MIN}$ metrics. The ranking columns indicate the order of methods according to the respective metric mean.}
\label{tab:results_third_experiment}
\end{table}

\begin{figure}[!ht]
	\centering
	\subfloat[]{\includegraphics[width=0.85\textwidth]{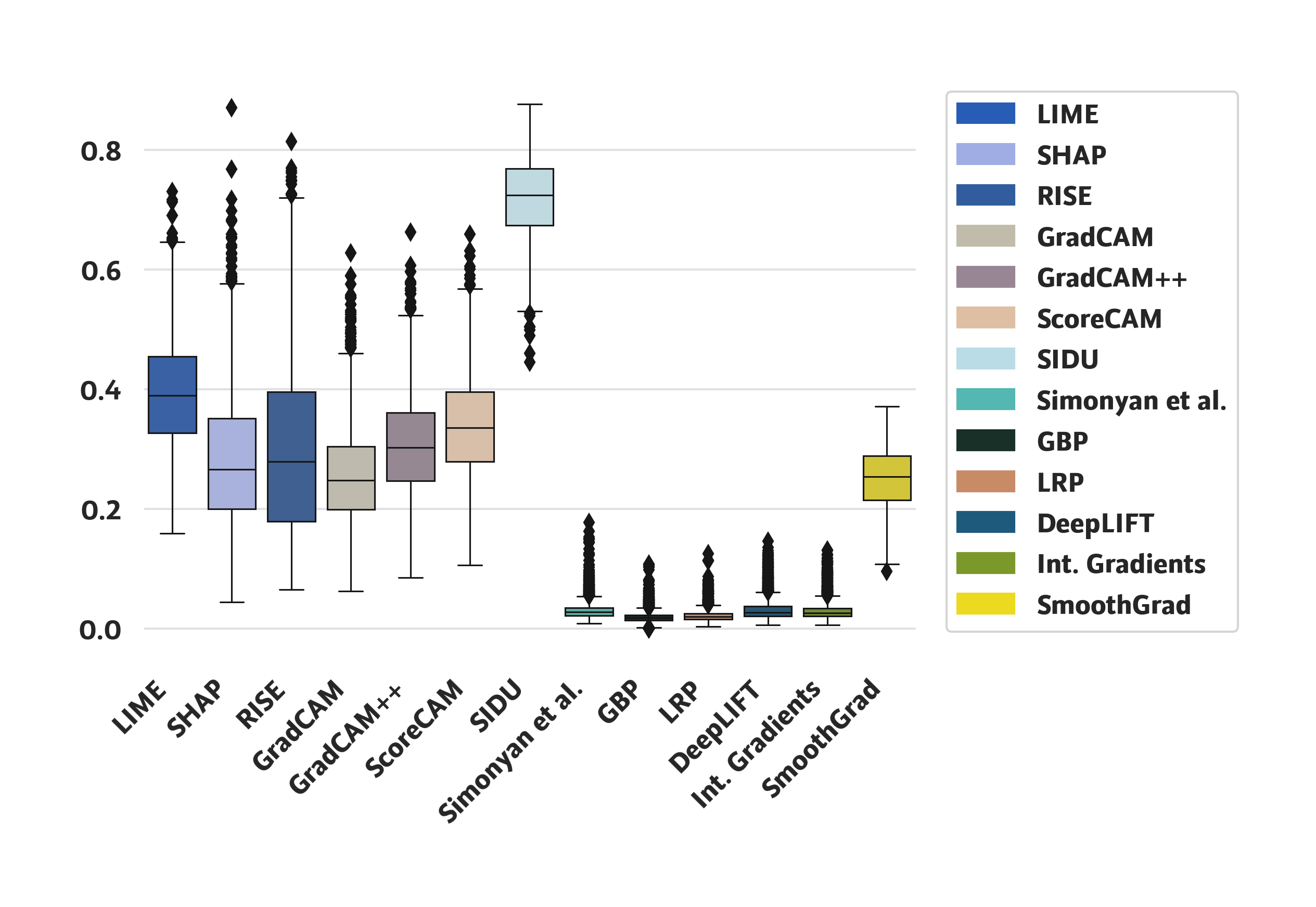}%
	\label{subfigure:v3_bp_emd}}
	\hfil
	\subfloat[]{\includegraphics[width=0.85\textwidth]{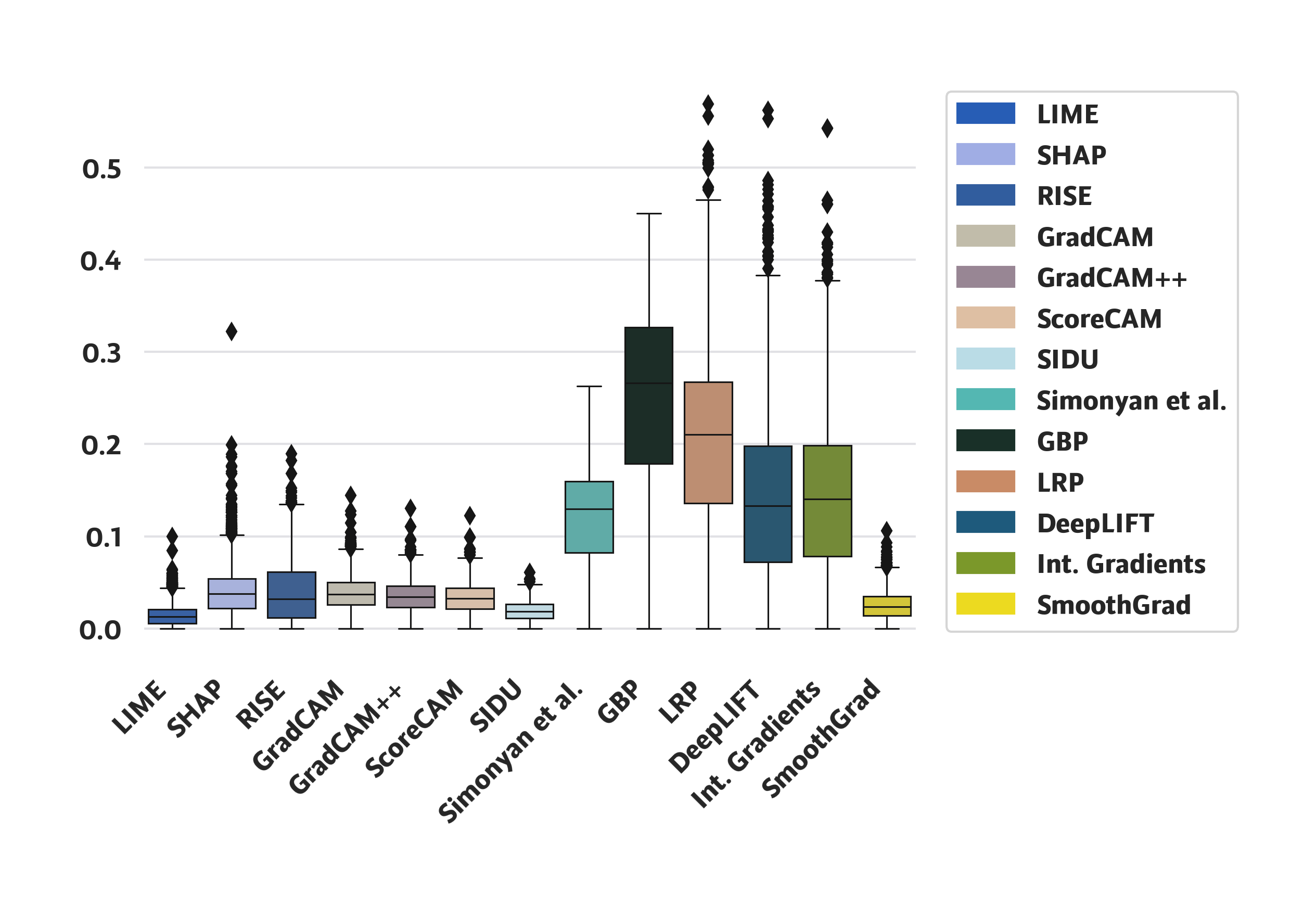}}%
	\label{subfigure:v3_bp_min}
	\caption{Box-plot for the $\mli{EMD}$ (a), and $\mli{MIN}$ (b) metrics obtained in the third experiment.}
	\label{figure:v3_bp}
\end{figure}

From the table \ref{tab:results_third_experiment} we can see that the best method is GBP~\cite{springenberg2014striving}. Once again, the best methods are the ones based on back propagating the output information to the input data, according both to the mean of the metrics and the standard deviation. Nevertheless, we can see that the differences between CAM methods and occlusion-based is lower than in the previous experiments. We hypothesize that the reason behind the improvements of the sensitivity based methods is due to the learned patterns of the neural network: because all images had different backgrounds, we believe that the model had learned to ignore those and for this reason, it is harder to generate OOD samples. This hypothesis can be supported by the fact that the $\mli{MIN}$ metric for backpropagation methods had significantly better results than in the previous experiment, meaning that there is less noise in the images, and at the same time, that the background is ignored.

In Figure \ref{figure:cmp_v3}, an example of the results obtained for the third experiment can be seen, with explanations compatible with the metrics values.
\FloatBarrier
\section{Conclusion}~\label{sec:conclusion}

In this study we compared objectively thirteen state-of-art XAI methods. To be able to do this comparisson we generated three new datasets with GT for the explanations, following the proposed methodology by Miró-Nicolau \emph{et al.}~\cite{miro2023novel}.

The experimental results demonstrated that the methods based on backpropagation of the output to the input data \cite{bach2015pixel, simonyan2014very, sundararajan2017axiomatic, springenberg2014striving, shrikumar2017learning} produced explanations with higher fidelity in all experiments. However, these methods tend to generate noise when the non-important areas have high activation values, as observed with the slow values of the $\mli{MIN}$ metric. The only exception was the work Smilkov \emph{et al.}\cite{smilkov2017smoothgrad} that had bad results, we considered that the cause of these results are the generation of OOD samples. 

Occlusion methods \cite{ribeiro2016should, Petsiuk2018, lundberg_unified_2017} performed poorly in all experiments. These methods bad results are caused, as discussed by Montavon~\emph{et al.}~\cite{montavon2018methods} because they indicated how a change in the input affects the output, and the real goal of an explanation is not to know how to change the input to obtain a different result, but to understand what makes an input produce a particular output. The CAM methods \cite{selvaraju2017grad, chattopadhay2018grad}, also had bad results cause by the misalignment of the explanations that generate the UpSampling operation, as studied and demonstrated by Xia \emph{et al.}~\cite{xia2021receptive}. Finally, the methods that combine these approaches \cite{muddamsetty2022visual, wang2020score} are susceptible of having both issues, and for this reason also have bad results. 

As future work, and after identifying the backpropagation methods as the models with the best fidelity to the backbone model, we consider necessary to improve these methods, in particular, aiming to reduce the amount of noise generated. This noise generates visible artefacts to the human user, and can affect the trust of the human user to the overall explanation.

\section{Declaration of competing interest}

The authors declare that they have no known competing financial interests or personal relationships that could have influenced the work reported in this study.


\section{Funding}
Project PID2019-104829RA-I00 “EXPLainable Artificial INtelligence systems for health and well-beING (EXPLAINING)” funded by \\ MCIN/AEI/10.13039/501100011033. Miquel Miró-Nicolau benefited from the fellowship FPI\_035\_2020 from Govern de les Illes Balears.

\bibliographystyle{abbrvnat} 
\bibliography{references}





\end{document}